\documentclass[aip,amsmath,amssymb,reprint]{revtex4-1}

\usepackage{graphicx}
\usepackage{dcolumn}
\usepackage{bm}

\usepackage[utf8]{inputenc}
\usepackage[T1]{fontenc}
\usepackage{mathptmx}
\usepackage{etoolbox}
\usepackage[usenames]{color}
\usepackage{comment}
\usepackage[normalem]{ulem} 

\usepackage{verbatim}

\makeatletter
\def\@email#1#2{%
 \endgroup
 \patchcmd{\titleblock@produce}
  {\frontmatter@RRAPformat}
  {\frontmatter@RRAPformat{\produce@RRAP{*#1\href{mailto:#2}{#2}}}\frontmatter@RRAPformat}
  {}{}
}%
\makeatother
\begin{document}

\preprint{AIP/123-QED}

\title[Dynamics-informed reservoir computing]{Dynamics-Informed Reservoir Computing with Visibility Graphs}
\author{Charlotte Geier}%
    \affiliation{Dynamics Group, Department of Mechanical Engineering, Hamburg University of Technology, Germany}
    \email{charlotte.geier@tuhh.de}

\author{Rasha Shanaz}%
\affiliation{Department of Physics, Bharathidasan University, Tiruchirappalli, India}
    
\author{Merten Stender}
    \affiliation{Chair of Cyber-Physical Systems in Mechanical Engineering, Technische Universität Berlin, Germany}

\date{\today}

\begin{abstract} 
Accurate prediction of complex and nonlinear time series remains a challenging problem across engineering and scientific disciplines. Reservoir computing (RC) offers a computationally efficient alternative to traditional deep learning by training only the read-out layer while employing a randomly structured and fixed reservoir network. Despite its advantages, the largely random reservoir graph architecture often results in suboptimal and oversized networks with poorly understood dynamics. Addressing this issue, we propose a novel Dynamics-Informed Reservoir Computing (DyRC) framework that systematically infers the reservoir network structure directly from the input training sequence. This work proposes to employ the visibility graph (VG) technique, which converts time series data into networks by representing measurement points as nodes linked by mutual visibility. The reservoir network is constructed by directly adopting the VG network from a training data sequence, leveraging the parameter-free visibility graph approach to avoid expensive hyperparameter tuning. This process results in a reservoir that is directly informed by the specific dynamics of the prediction task under study. We assess the DyRC-VG method through prediction tasks involving the canonical nonlinear Duffing oscillator, evaluating prediction accuracy and consistency. Compared to an Erd\H{o}s-R\'enyi (ER) graph of the same size, spectral radius, and fixed density, we observe higher prediction quality and more consistent performance over repeated implementations in the DyRC-VG. An ER graph with density matched to the DyRC-VG can in some conditions outperform both approaches.
\end{abstract} 

\maketitle
\begin{quotation}
As reservoir computing gains popularity for time series forecasting tasks in complex systems, the search for an optimal design of the reservoir structure becomes more important. Today, a generic structure-function relationship in the context of reservoir computing is unknown. This article proposes replacing the random reservoir setup, which often requires expensive hyperparameter tuning, with a deterministic, dynamics-driven one. The dynamics-informed reservoir computing approach (DyRC-VG) translates the training time series into a visibility graph whose structure serves as the reservoir, thereby linking the intrinsic dynamics of the target system with the reservoir structure. 
\end{quotation}


\section{\label{sec:intro}INTRODUCTION}
Reservoir computing has become a popular form of machine learning for applications such as time series forecasting \cite{Carroll.2019, Zhang.2021, Yadav.2024}. In contrast to deep neural networks, where the network structure is set up in layers, the reservoir has a random structure that generates a high-dimensional latent representation of the inputs \cite{Nakajima.2021, Tanaka.2019}. An input layer distributes the input across the reservoir, and a readout layer is trained to map the reservoir dynamics to the target dynamics. Since the weights of the readout layer are the only ones that are adjusted during the training phase, reservoir computers (RC) are associated with low computational effort compared to deep learning approaches. RCs are efficient in scenarios involving smaller datasets compared to data-hungry deep learning models \cite{Nishioka.2024, Yadav.2024}. At the same time, RCs generalize well since their performance relies on the fixed dynamic properties of the reservoir to capture temporal patterns in the data \cite{Ma.2023, Yan.2024}. Therefore, RCs are well-suited to process sequential data, such as time series.

Currently, the structure of RC based on echo state networks (ESN) \cite{Jaeger.2001} is mostly defined at random \cite{Nakajima.2021, Tanaka.2019}, classically as Erd\H{o}s-R\'enyi (ER) graphs, with no formalized or structured approach to guide the setup \cite{Dale.2019}. Several works study the impact of specific topological properties on RC performance. For example, Dale et al \cite{Dale.2019}examine the effect of reservoir connectivity on its performance by adding random connections to regular structures such as lattices and rings. The authors find that reservoir performance increases with a small percentage of additional edges. Dale et al \cite{Dale.2021} find that smaller networks with more complex connectivity structure can perform as well as larger networks with simpler structure, underlining the importance of topology over pure reservoir size. Rather et al\cite{Rathor.2025} show that symmetry or lack thereof is an important factor in reservoir performance. Despite these efforts,  specific structure-function relationships remain poorly understood \cite{Yadav.2025}, which hinders the development of a unified approach to defining an optimal structure for a given task. The reservoir computer will therefore often be unnecessarily large. Several methods are being studied to develop a deterministic way of finding an `optimal' reservoir. For example, Yadav et al. generate task-optimized minimal reservoir structures through a performance-driven network evolution scheme \cite{Yadav.2025}. Several hybrid approaches that combine knowledge-based models with reservoir computers exist \cite{Pathak.2018, Shannon.2025}. For example, Köster et al. propose a data-informed-reservoir computing (DIRC) approach that combines a knowledge-based sparse identification of nonlinear dynamics (SINDy) \cite{Brunton.2016} model with a reservoir computer to increase the forecasting horizon and reduce the cost of hyperparameter tuning \cite{Koster.2023}.

Motivated by the above considerations, we expect that specific reservoir structures exhibit superior information processing capabilities compared to other structures. One way to inform the RC's structure is to align it with the dynamical process encountered in the prediction task, for example, by translating time series data from the prediction task into a graph structure of the reservoir. By constructing reservoirs based on adjacency matrices derived from visibility graphs \cite{Lacasa.2008}, our dynamics-informed reservoir computer (DyRC-VG) establishes a direct link between the system dynamics of the prediction task and the architecture of the RC, thus generating an automated method of adapting the graph structure to a given problem. The visibility graph is a parameter-free approach that measures the convexity of time series segments between the data points \cite{Lacasa.2008}. Visibility graphs have been shown to encode important structural information on the dynamics of the underlying system \cite{Lacasa.2008, Luque.2009, Lacasa.2010, Stephen.2015, Ni.2009}, such as the type of dynamics \cite{Lacasa.2008, Lacasa.2010, Pierini.2012, Lacasa.2014, Ravetti.2014} and process reversibility \cite{Lacasa.2012, Donges.2013tts}. Therefore, the method represents an interesting candidate for introducing information on the underlying system into the setup of the RC's structure. Our work is thus a contribution to the larger field of research \cite{Yadav.2025} that tries to link structure and function in information-processing graphs. 

In this work, we study how the structure arising from visibility graphs affects the information processing capabilities of RCs. We employ different variants of leveraging visibility graphs for setting up the reservoir computer, seeking to determine the effect of a dynamics-informed reservoir structure on model accuracy and robustness. The results show that a dynamics-informed reservoir may indeed outperform a purely random one. 

The remainder of this work is structured as follows. This introduction is followed by the presentation of the proposed method in Section \ref{sec:method}. Section \ref{sec:reults} presents results from numerical studies, illustrating the performance of visibility-graph-based reservoirs compared to randomly set up structures. The paper is completed by a conclusion and outlook in Section \ref{sec:concl}. 

\section{\label{sec:method}DYNAMICS-INFORMED RESERVOIR COMPUTING}
\begin{figure}[!htb]%
    \centering
    \includegraphics[width=0.4\textwidth, trim= 0 10 0 6, clip]{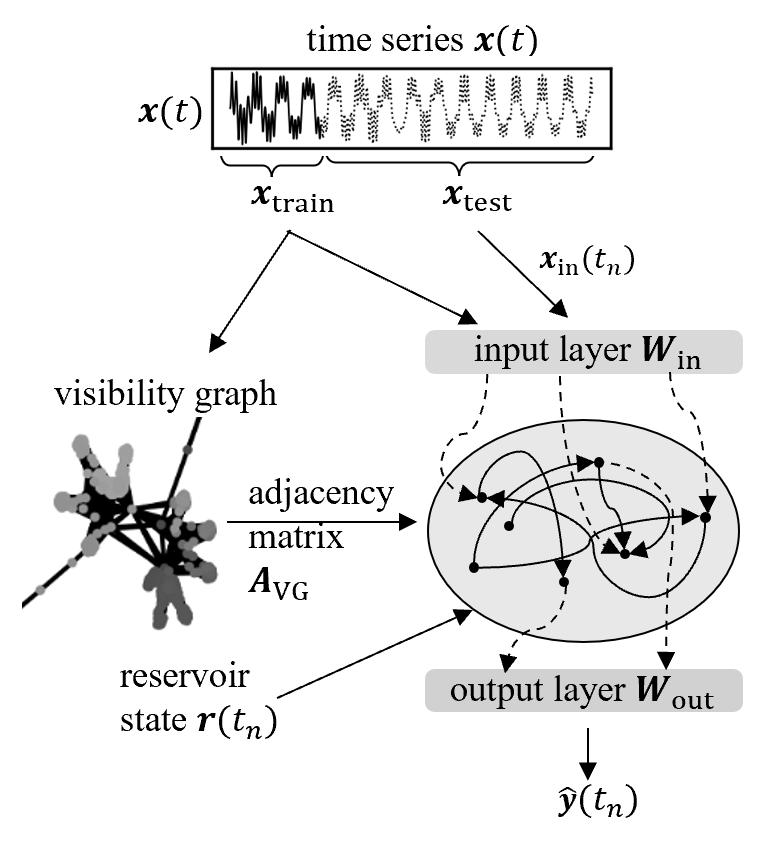}
    \caption{A dynamics-informed reservoir computer (DyRC-VG). Time series data from a dynamical system is translated into a network in the form of a visibility graph. The structure of the graph is used to construct the reservoir computer. The reservoir computer is trained as a predictive model for the dynamical system.}
    \label{fig:figure_1}
\end{figure}

In this work, we propose a novel approach to structuring the reservoir network as presented in Figure \ref{fig:figure_1}. In the first step, the training time series data $\mathbf{x}(t)$ is mapped to the corresponding visibility graph with adjacency matrix $\mathbf{A}_\mathrm{VG}$. After scaling to a desired spectral radius, that adjacency matrix is set as a reservoir network for an RC model. Then, the reservoir computer is trained on the training sequence and validated on a test sequence. The prediction performance is compared with that of the classical reservoir computer constructed from an ER graph with the same size, density, and comparable spectral radius as the dynamics-informed matrix $\mathbf{A}_\mathrm{VG}$. 

\subsection{\label{subsec:rc}Reservoir computing}

The reservoir computing framework is derived from recurrent neural networks. In contrast to deep learning approaches, an RC consists of only three layers, namely an input layer $\mathbf{W}_\mathrm{in}$, a reservoir layer $\mathbf{A}$, and a readout layer $\mathbf{W}_\mathrm{out}$, with the readout layer being the only one that is adjusted during the training phase. 
As a dynamics input signal $\mathbf{x}_{\mathrm{in}}(t_n)$ excites the reservoir at time $t_n$, its internal dynamics $\mathbf{r}(t_n)$ at time $t_{n+1}$ are updated according to 
\begin{align}
    \mathbf{r}(t_{n+1}) &= (1-\alpha) \mathbf{r}(t_n) + \alpha f(\mathbf{A}\mathbf{r}(t_n) + \mathbf{W}_\mathrm{in} \mathbf{x}_\mathrm{in}(t_n)) \\
    \mathbf{y}(t_n) &= \mathbf{W}_\mathrm{out} \mathbf{r}(t_n), \label{eq:y_t}
\end{align}
where $\mathbf{r}(t_n) \in \mathbb{R}^{N}$ is the reservoir state vector of a reservoir with $N$ nodes at given time $t_n$ and $\mathbf{x}_\mathrm{in}(t_n) \in \mathbb{R}^m$ constitutes the external dynamic input, where $m$ denotes the number of input variables. The input layer $\mathbf{W}_\mathrm{in} \mathbb{R}^{N \times m}$ determines at which nodes the input data is fed into the reservoir, the adjacency matrix $\mathbf{A} \in \mathbb{R}^{N \times N}$ represents the internal connections between reservoir units, $\alpha \in [0,1]$ encodes the leakage of past reservoir information leaked over time, and $f$ nonlinear activation function for each node. The readout layer $\mathbf{W}_\mathrm{out} \in \mathbb{R}^{m \times N}$ maps reservoir states $\mathbf{r}(t_n)$ to the estimated output dynamics $\mathbf{y}(t_n)$. In contrast to conventional deep learning approaches, only the weights of the readout layer $\mathbf{W}_\mathrm{out}$ are adjusted during training, commonly using Ridge regression on the set of reservoir states $\mathbf{R} \in \mathbb{R}^{N \times T}$ linearly combined to the target sequence $\mathbf{Y} \in \mathbb{R}^{m \times T}$, where $T$ is the total time series length considered. 

In this work, the RC is trained to predict the states $\mathbf{y}(t_n) = [q(t_n), \dot{q}(t_n)]$ of a forced nonlinear Duffing oscillator system (see Appendix \ref{app:model}) at time $t_n$ from the states at time $t_{n-1}$ and the forcing at time $t_n$, such that the input is given by $\mathbf{x}_\mathrm{in}(t_n)=[q(t_{n-1}), \dot{q}(t_{n-1}, g(t_n)]$. During the training phase, the model is given both the input and output of a training section of the dynamics. In deployment, a different section of forcing is given to the model, which then predicts the Duffing dynamics $\mathbf{\hat{y}}(t_n)$. 

The structure of the random reservoir is constructed as an Erd\H{o}s-R\'enyi (ER) graph \cite{Erdos.1959} with density $\rho=0.1$ and leakage rate $\alpha=0.5$. Generally, two main principles guide the design of the reservoir structure $\mathbf{A}$: The assumption that more nodes will make the computer more capable, and the recent finding that a sparser RC is more efficient than a dense one \cite{Yadav.2025}. In the following, we link the structure of the RC to the prediction task's dynamics by using visibility graphs. The reservoir computations in this work are performed using the \texttt{pyReCo} library \cite{pyreco} in Python. The parameter settings for the RC can be found in Appendix \ref{app:rc_param}.  

\subsection{\label{subsec:vis_graph}Visibility graphs}
Introduced by Lacasa et al \cite{Lacasa.2008}, visibility graphs (VG) represent a hyperparameter-free method for translating time series data into a network structure. The approach is popular due to its simplicity \cite{Donner.2012vga}, for example, to classify different types of dynamics \cite{Luque.2009, Lacasa.2010}. Each data point in the time series becomes a node in the graph. Nodes are connected based on the mutual visibility of the respective data values defined by the convexity of the time series segment between them. Two data points $(x_i, t_i)$ and $(x_j, t_j)$ are thus connected iff 
\begin{equation}
    x_l < x_j + (x_i - x_j) \frac{t_j-t_i}{t_j-t_l}
\end{equation}
for any other point $(x_l, t_l)$ of data between them. The resulting graph is an undirected, fully connected graph described by the symmetric adjacency matrix $\mathbf{A}_\mathrm{VG}$.
While the temporal information is lost in the translation from time series to graph, the graph inherits structural properties from the time series. For example, each convex section of the time series translates into a node cluster, with maxima as hubs to connect them. Consequently, random time series tend to form random graphs, while fractal time series tend to translate into scale-free networks \cite{Lacasa.2008}.

In the DyRC-VG approach, the VG is generated from a section of the position variable $q$ of the Duffing system. For example,  a reservoir of size $N=100$ requires a 100-time-step section of the training data. To avoid introducing bias from the specific sample picked, we chose 100 different sections uniformly distributed within the training dataset. 

To integrate more dynamic information into the DyRC-VG while keeping the number of reservoir nodes low, a variant, DyRC-VG 16, is implemented. In the DyRC-VG 16 version, only every 16th data point of the training time series is used to compute the VG, allowing the use of more dynamic information in the same reservoir size. Care was taken to avoid mixing information used to inform and train the reservoir with the test data.

After computation of the VG, every graph is normalized to a spectral radius $\nu=0.9$, comparable to that of the random network. VG computation is performed using the NetworkX \cite{networkx} package in Python.

\section{\label{sec:reults}RESULTS}

To assess the performance of the DyRC-VG approach, five different scenarios are studied:
\begin{description}
    \item[ER] a `standard' random reservoir with ER-structure, density $\rho_\mathrm{rand}= 0.1$, and spectral radius $\nu = 0.9$ as the baseline reservoir,
    \item[DyRC-VG] the DyRC-VG with a visibility graph reservoir is normalized to spectral radius $\nu = 0.9$, where the density $\rho_\mathrm{VG}$ is given by the VG, 
    \item[ER $\rho$(VG)] a random reservoir with ER-structure, spectral radius $\nu = 0.9$, and density $\rho_\mathrm{VG}$ comparable to the DyRC-VG,
    \item[DyRC-VG 16] the DyRC-VG 16, where the VG is computed from a down-sampled time series section, and the reservoir is normalized to spectral radius $\nu = 0.9$, where the density $\rho_\mathrm{VG}$ is given by the VG , and
    \item[ER $\rho$(VG16)] a random reservoir with ER-structure, spectral radius $\nu = 0.9$, and density $\rho_\mathrm{VG}$ comparable to the DyRC-VG 16.
\end{description}

The performance of each approach is quantified using the mean absolute error $MAE = 1/n \sum_{i=1}^n |\hat{\mathbf{y}}_i-\mathbf{y}_{\mathrm{test},i}|$ between the true system states $\mathbf{y}_\mathrm{test}$ and the predicted dynamics $\hat{\mathbf{y}}$. 

\begin{figure}
    \centering
    \includegraphics[width=0.45\textwidth]{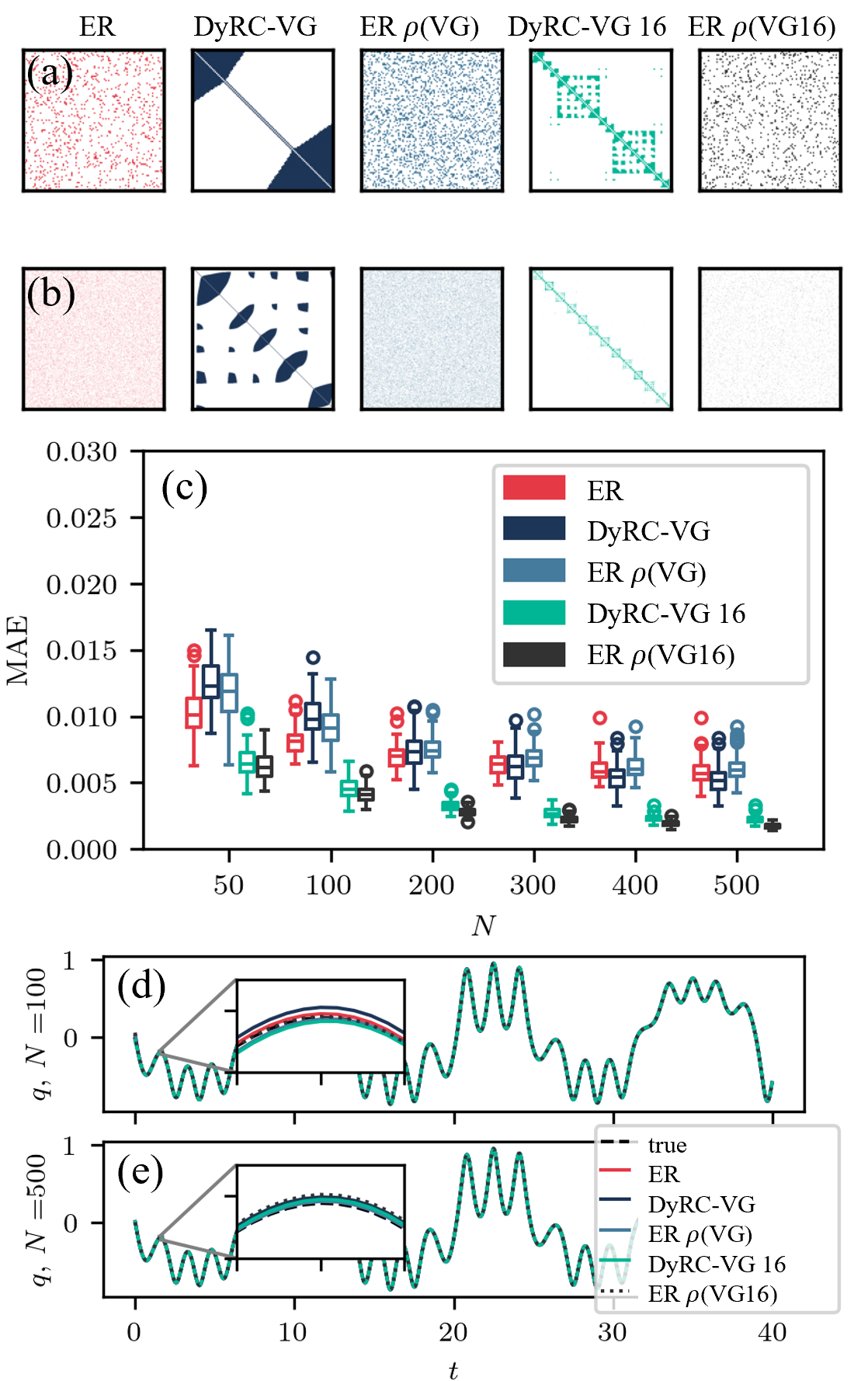}
    \caption{Performance of DyRC-VG and DyRC-VG 16 in comparison with the ER reservoir structure. (a,b)Exemplary matrices of the five settings, an ER reservoir (red), the DyRC-VG (dark blue), the ER $\rho$(VG) with comparable density (blue), the DyRC-VG 16 (green) from a coarse-grained time series, and the ER $\rho$(VG16) with matched density, are shown for (a) $N=100$ (top row) and (b) $N=500$ (second row) nodes. (c) The performance of each approach is measured in terms of its predictive capability as the MAE between true and predicted time series, shown in the middle panel for different reservoir sizes with 100 implementations each. (d,e) The bottom panels illustrate exemplary time series prediction for the Duffing position variable $q$ and two reservoir sizes $N=[100,500]$, in the same color code.}
    \label{fig:figure_2}
\end{figure}

The central panel of Figure \ref{fig:figure_2} shows the performance of all five variants for different reservoir sizes 
$N=[50,100,200,300,400,500]$, along with exemplary adjacency matrices and prediction time series. For each setting, 100 different implementations are evaluated, meaning that 100 different random (ER) matrices with similar density and spectral radius are computed for the three random scenarios, while each VG implementation uses a different time series section, resulting in a range of densities. From left to right, the top panels in Figure \ref{fig:figure_2} show the adjacency matrices for a `standard' random (ER) reservoir setup with a density of $\rho=0.1$ in red, a DyRC-VG matrix without down-sampling of the time series in dark blue, an ER $\rho$(VG) implementation in blue, a DyRC-VG 16 matrix in green, and a an ER $\rho$(VG16) in black. The first and second rows depict these matrices for reservoir sizes $N=100$ and $N=500$, respectively. The bottom panels illustrate the true and predicted time series for the position variable of the Duffing oscillator. The color code is the same as for the matrices and box plot above, the upper panel shows the results for reservoir size $N=100$, the bottom panel shows the results for size $N=500$.

The metrics in Figure \ref{fig:figure_2} indicate that the DyRC-VG 16 ER $\rho$(VG16) approaches perform better than the other three approaches both in terms of mean error and in terms of variability in the prediction results. The sparse ER structure outperforms the VG structure slightly. The predictions of the DyRC-VG approach appear worse than those of the ER counterparts for smaller reservoir sizes, but seem slightly better for larger reservoirs. Overall, larger reservoirs perform better in the prediction task at hand. 

\begin{figure*}
    \centering
    \includegraphics[width=0.9\textwidth]{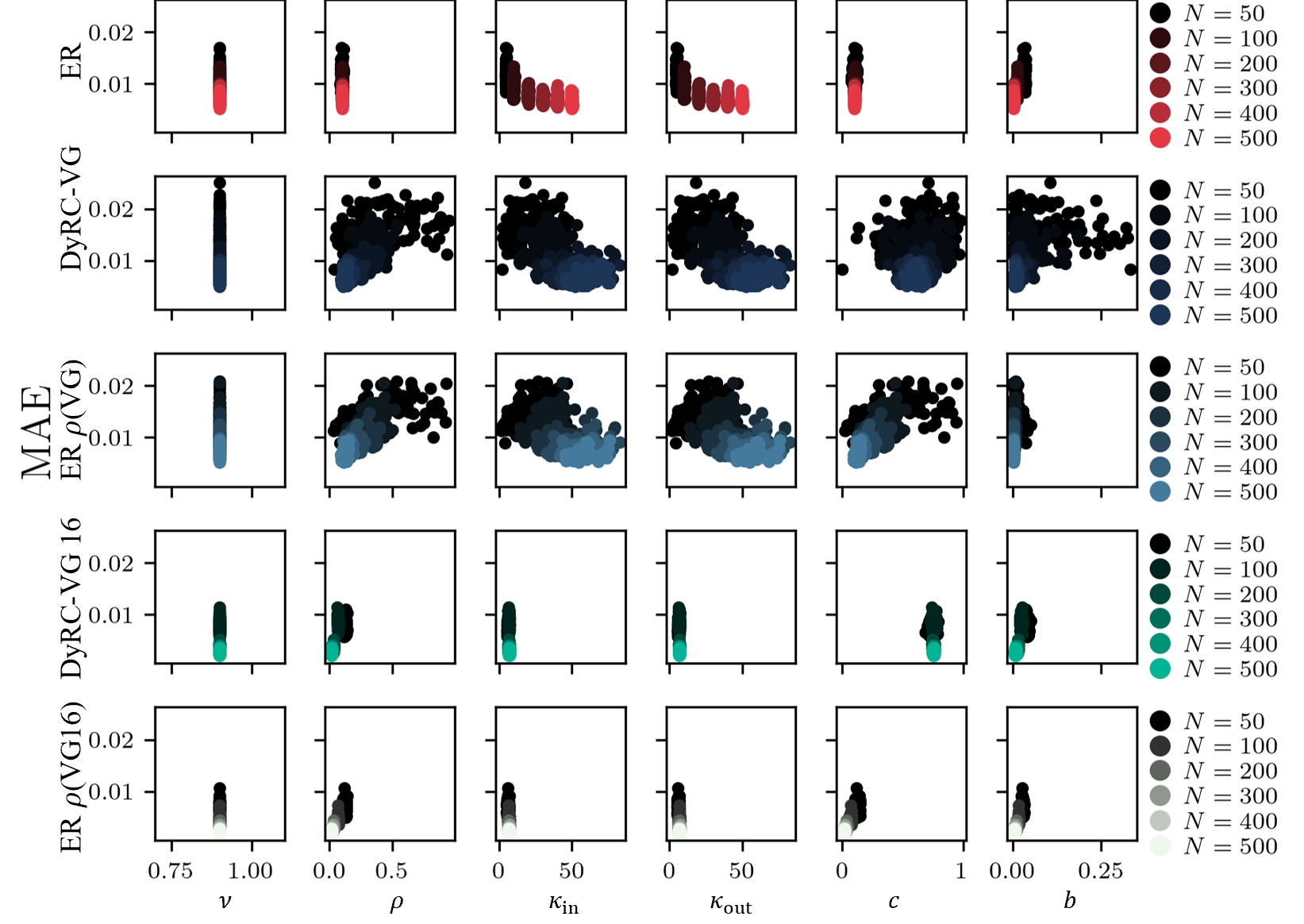}
    \caption{RC performance in relation to network metrics. Each row depicts the MAE metric for the five} versions, DyRC-VG, ER $\rho$(VG), DyRC-VG 16, and ER $\rho$(VG16) over a different network metric in each column. The spectral radius $\nu$ is given by the RC setup and acts as a control parameter. The density $\rho$ of the random RC is defined as 0.1, while arising from the time series data in both VG versions. The random dense RC has a density oriented with that of the DyRC. The remaining network metrics, average in-degree $\kappa_\mathrm{in}$, average out-degree $\kappa_\mathrm{out}$, global clustering coefficient $c$, and average betweenness centrality $b$, arise from the RC structure. For the definition of the metrics, see Appendix \ref{app:nw_metrics}. Color gradients represent the number of nodes $N$ in the reservoir, from $N=50$ in black to $N=500$ in the respective colors.
    \label{fig:figure_3}
\end{figure*}

The observations made in Figure \ref{fig:figure_2} raise a number of questions: Why is the variability of the results with the DyRC-VG 16 and the ER $\rho$(VG16) significantly lower than that of the other approaches? And: Can network metrics determine the success of the dynamics-informed approach? In order to answer these questions, Figure \ref{fig:figure_3} presents a more detailed insight. For each variant, the MAE is plotted over the different network metrics spectral radius $\nu$, density $\rho$, average in-degree $\kappa_\mathrm{in}$, average out-degree $\kappa_\mathrm{out}$, degree centrality $c$ and betweenness centrality $b$. The color code is that same as in Figure \ref{fig:figure_2}, with the ER $\rho$(VG16) shown in black/mint. For a detailed description of the metrics, see Appendix \ref{app:nw_metrics}. Within each panel, the results are grouped by the different network sizes $N$, represented by several color shades. 
By definition, the spectral radius $\nu=0.9$ is constant for all implementations. The density $\rho_\mathrm{ER}=0.1$ of the ER graph is also prefixed. The density varies in the visibility graph implementations as this property arises directly from the computation, and the densities of the ER $\rho$(VG) and ER $\rho$(VG16) graphs follow that of the DyRC-VG and DyRC-VG 16. Compared to the DyRC-VG, the density of the DyRC-VG 16 appears less varied over the number of implementations. Presumably, this effect stems from the longer time series section involved in the DyRC-VG 16 computation, which results in more homogeneous matrices across different implementations. For larger reservoir sizes, the density of the DyRC-VG 16 adjacency matrix drops below that of the ER graph with fixed density, underlining the common intuition that sparse reservoirs tend to perform better. 
For all remaining metrics, the DyRC-VG and, correspondingly, the sense ER approach, exhibit considerable variability. This variability decreases as the number of nodes in the reservoir increases, which is to be expected, since a longer section of the time series segment captures multiple periods rather than just one or a fraction of one, thereby averaging out fluctuations. This phenomenon also explains the significantly lower variance in the DyRC-VG 16, and, correspondingly, in the ER $\rho$(VG16), thus effectively answering the first question posed.
The average degrees $\kappa_\mathrm{in}$ and  $\kappa_\mathrm{out}$ follow from the densities in combination with the reservoir sizes: In the ER setting, a constant density is maintained while increasing the number of nodes in the graph, thus increasing the average in- and out-degree, while in the DyRC-VG 16 setting, the density decreases with the rising umber of nodes, resulting in a near-constant average degree. Further studies, including the degree distribution of the respective graphs, might yield additional information.
The DyRC-VG 16 demonstrates a much higher clustering coefficient compared to its random counterpart. This difference is also evident in the matrices presented in Figure \ref{fig:figure_2}. 
For certain cases (e.g., N=500), the adjacency matrix exhibits structural properties resembling a simple cycle reservoir (SCR), whose directed ring topology offers universality for fading-memory filters \cite{li2024}. However, in DyRC-VG, these features emerge naturally from the visibility graph of the input dynamics, rather than being imposed as a fixed structure.
%
To capture the slowest dynamics, we followed the intuition that the VG window size should be at least as long as the system's longest characteristic timescale, $T_\mathrm{char}$. For $N>100$, the VG window exceeds $T_\mathrm{char}$ of the Duffing system ($\approx 2.93t$), but even smaller DyRC-VG perform well, likely because the VG edges encode multiscale correlations that capture the slower dynamics without covering the full period.

%
Attempting to answer the second question, the most important network metric appears to be either clustering or density; further studies are needed to determine which, or perhaps both, factor is the crucial one. 
Similar results have been observed for the different chaotic Duffing system implementations as detailed in Appendix \ref{app:model}, Figure \ref{fig:figure_4}. For completeness, our approach was also evaluated on the canonical Lorenz and Mackey-Glass systems. Results, presented in Appendix \ref{app:supp_benchmark}, exhibit comparable performance across all DyRC-VG variants.

\section{\label{sec:concl}CONCLUSION}
Dynamics-informed reservoir computers with visibility graphs (DyRC-VG) encode dynamical properties in their reservoir structure. Compared to random ER configurations, this task-tailored approach exhibits superior predictive capabilities when applied to different configurations of a Duffing oscillator, if a sufficient amount of information is incorporated into the structure. In terms of practical application, several cycles of a time series should be embedded into the visibility graph to get consistent results. 
Overall, the integration of structure and function in this computational setting seems to be a promising avenue for investigation. Future research may explore alternative strategies for embedding dynamical information into reservoir computing frameworks, including visibility graph variants \cite{Bezsudnov.2014,Zhou.2012}, and along with a more detailed analysis of relevant network metrics. We hope to stimulate further investigations into the relationship between function and structure in reservoir computing for complex dynamical systems.

\begin{acknowledgments}
\vspace{-3mm}
This work was supported by the Deutsche Forschungsgemeinschaft (DFG, German Research Foundation) under the Special Priority Program (SPP 2353: Daring More Intelligence—Design Assistants in Mechanics and Dynamics, project number 501847579). CG is thankful to the DFG for support through project number 510246309. RS thanks Prof P. Muruganandam, their PhD supervisor, for his support, guidance, and constructive input in the formative phases of this work.
\end{acknowledgments}
\vspace{-3mm}

\section*{Conflict of Interest Statement}
\vspace{-3mm}

The authors have no conflicts to disclose.
\vspace{-3mm}

\section*{Author Contributions}
\vspace{-3mm}
\textbf{CG:} Conceptualization (supporting); methodology (equal); software (equal); writing - original draft (lead); writing - review and editing (lead). \textbf{RS:} Conceptualization (supporting); methodology (supporting); software (supporting); writing - review and editing (equal). \textbf{MS:} Conceptualization (lead); methodology (equal); software (equal); supervision (lead); writing - review and editing (equal).
\vspace{-3mm}

\section*{Data Availability Statement}
\vspace{-3mm}
The data and code that support the findings of this study are openly available in Zenodo at \href{https://doi.org/10.5281/zenodo.16410959}{https://doi.org/10.5281/zenodo.16410959}.
\textbf{}

\appendix

\section{Model system}\label{app:model}
\vspace{-3mm}
The results in this work are shown along three variants of the Duffing oscillator \cite{Kovacic.2011, Ueda.1991} with different parameter settings. The system equations are given by 
\begin{equation}
    \ddot{x} + d \dot{x} + kx + k_\mathrm{nl} x^3 = F \cos(\Omega t),
\end{equation}
where $\ddot{x}, \dot{x}$ and $x$ denote the acceleration, velocity and position variables, respectively. The damping is given by $d$, $k$ represents the linear spring stiffness, $k_\mathrm{nl}$ the nonlinear spring stiffness, $F$ the forcing amplitude, and $\Omega$ the forcing frequency. The parameters for each dataset are shown in Table \ref{tab:duffing_parameters}.
\begin{table}[h]
\vspace{-5mm}
    \caption{Duffing parameters}\label{tab:duffing_parameters}
    \vspace{0.5em}
    \begin{tabular}{l | l | l | l | l | l }
    \textbf{set} & \textbf{$d$} & \textbf{$k$}  & \textbf{$k_\mathrm{nl}$} & \textbf{$\Omega$} & \textbf{$F$}\\
    \hline
    1 & 0.02 & 1 & 5 & 8 & 0.5 \\
    2 & 0.1 & -1 & 0.25 & 2.5 & 2 \\
    3 &  0.1 & 1 & 2 & 35 & 2
    \end{tabular}
\vspace{-10mm}
\end{table}

\begin{figure}
    \centering
    \includegraphics[width=0.5\textwidth, , trim= 5 8 5 5, clip]{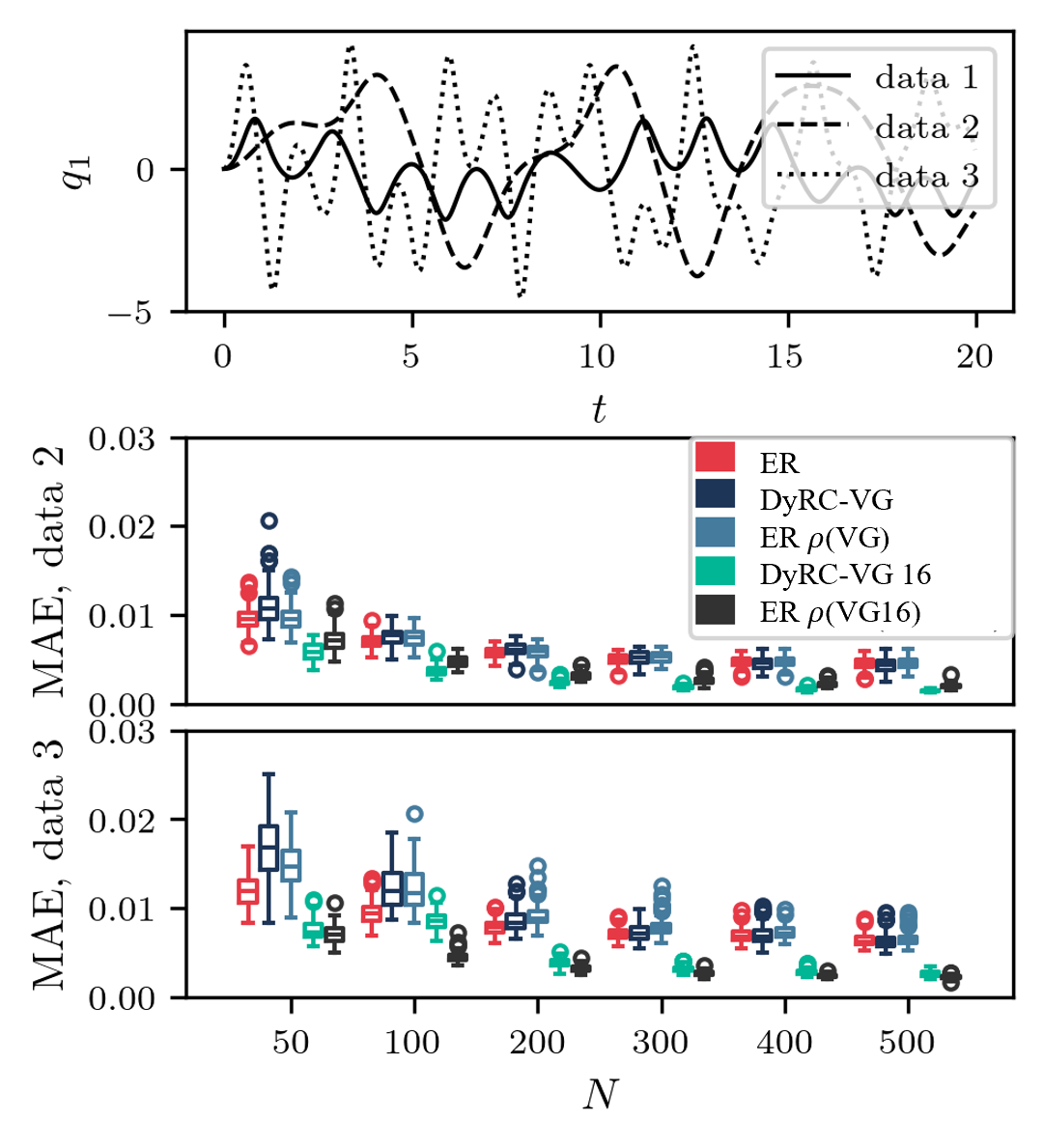}
    \caption{Different chaotic} Duffing versions and corresponding results.
    \label{fig:figure_4}
\end{figure}

\section{Reservoir parameters}\label{app:rc_param}
\vspace{-3mm}
Table \ref{tab:rc_params} shows the settings and parameter values for the reservoir computer in general, and the random ER setup in particular. 
\begin{table}[h]
\vspace{-5mm}
    \caption{Reservoir computer parameters}\label{tab:rc_params}
    \vspace{0.5em}
    \begin{tabular}{l | l | l }
    \multicolumn{3}{l}{General parameters} \\
    \hline
    input fraction & & 0.5 \\
    optimizer & & Ridge regression \\
    RC leakage & $\alpha$ & 0.5   \\
    nonlinear activation & $f$ & tanh \\
    \hline
    \multicolumn{3}{l}{ER reservoir setup} \\
    \hline
    density & $\rho$ & 0.1 \\
    \end{tabular}
    \vspace{-5mm} 
\end{table}

\section{Network metrics}\label{app:nw_metrics}
\vspace{-3mm}
This section gives a short overview of the network metrics used throughout this work. For more detailed information, visit the NetworkX or pyReCo documentation under \cite{networkx, pyreco}. The spectral radius $\nu = \max(\mathrm{eig}(\mathbf{A})$ is defined as the maximum eigenvalue of the the adjacency matrix. The density $\rho$ of of the RC is computed as 
\begin{equation}
    \rho = \frac{2E}{N(N-1)},
\end{equation}
where $N$ is the number of nodes and $E$ is the number of edges in the graph. The average in-degree $\kappa_\mathrm{in}$ and average out-degree $\kappa_\mathrm{out}$ are computed by counting the number of in- and out-edges of each node and taking the average of that value. The global clustering coefficient $c$ is defined by
\begin{equation}
    c = \frac{1}{N} \sum_{v \in G} \frac{2 T (v)}{\kappa(v) (\kappa(v)-1)},
\end{equation}
counting the number of triangles $T(v)$ through node $v$, dividing by the nodes degree $\kappa(v)$ and computing the average over all clustering coefficients $c_v$. Similarly, the average betweenness centrality $b$ is computed as the 
\begin{equation}
    b = \frac{1}{N} \sum_{v \in G} \sum_{s,w \in G} \frac{\sigma(s, w |v)}{\sigma(s, w)},
\end{equation}
where $\sigma(s, w)$ is the number of shortest paths between nodes $s, w$, and $\sigma(s, w |v)$ is the count of those paths connecting through node $v$.

\section{Supplementary Benchmarks}\label{app:supp_benchmark}
\begin{figure}
    \centering
    \includegraphics[width=0.5\textwidth, trim= 5 5 0 5, clip]{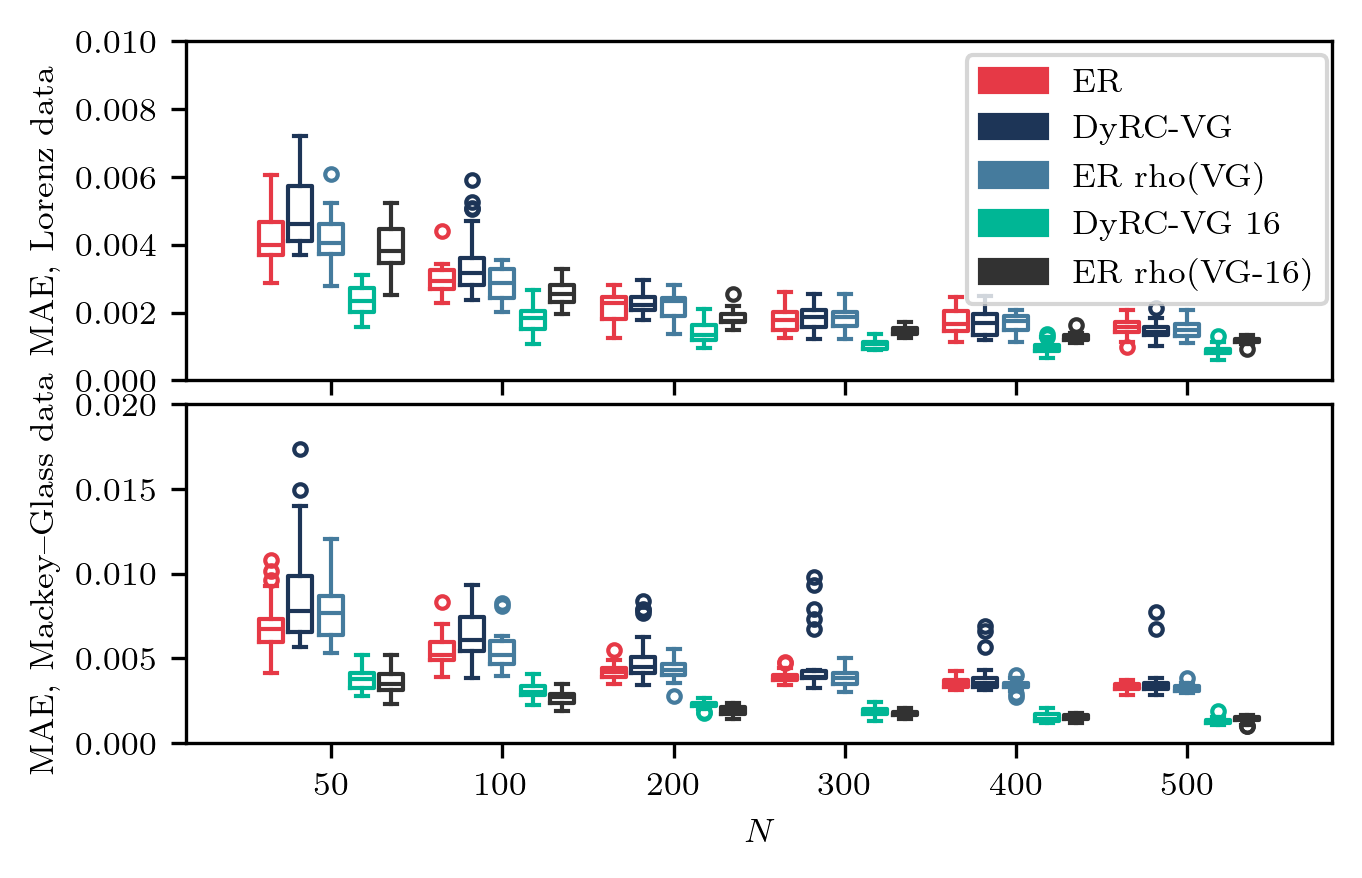}
    \caption{A comparison of performance of the various reservoir approaches for Lorenz and Mackey-Glass data}
    \label{fig:figure_5}
\end{figure}

\vspace{-3mm}
To assess the generality of the DyRC-VG method, two additional benchmark systems are evaluated, namely the Lorenz system, described by
 $$\dot x = \sigma(y-x),
    \dot y=x(\rho-z),
    \dot z = xy - \beta z $$
with standard chaotic parameter set $\sigma = 10.0, \rho = 28.0, \beta = 8/3$ and using $x(t)$ as the input signal and the Mackey-Glass system, described by the delay differential equations
$$
\dot x = \frac{\beta x(t-\tau)}{1+x(t-\tau)^n} - \gamma t
$$
with parameters set to $ \beta=0.2, \gamma=0.1, \tau=17, n=10$ for chaotic dynamics. 
The same pre-processing and visibility graph construction steps as in the Duffing system case are applied to both cases. For both systems, the observed performance is comparable to that seen in the Duffing benchmark, see Figure \ref{fig:figure_5}.

\section{References}
\vspace{-3mm}
\nocite{*}
\bibliography{references}

\end{document}